\def\year{2022}\relax
\documentclass[letterpaper]{article} 
\usepackage{aaai22}  
\usepackage{times}  
\usepackage{helvet}  
\usepackage{courier}  
\usepackage[hyphens]{url}  
\usepackage{graphicx} 
\usepackage{natbib}  
\usepackage{caption} 
\DeclareCaptionStyle{ruled}{labelfont=normalfont,labelsep=colon,strut=off} 
\usepackage{algorithm}

\usepackage{newfloat}
\usepackage{listings}
\floatstyle{ruled}
\newfloat{listing}{tb}{lst}{}
\floatname{listing}{Listing}
\usepackage{times}
\usepackage{latexsym}

\usepackage[T1]{fontenc}

\usepackage{url}            
\usepackage{booktabs}       
\usepackage{amsfonts}       
\usepackage{nicefrac}       
\usepackage{microtype}      

\usepackage{algorithmicx}
\usepackage{algorithm}
\usepackage{algpseudocode}
\usepackage{amsmath} 
\usepackage{color}
\usepackage{graphicx}
\usepackage{subcaption}
\usepackage{multirow}
\usepackage{wrapfig}

\newcommand{\gv}{{\boldsymbol g}}

\newcommand{\xv}{{\boldsymbol x}}
\newcommand{\yv}{{\boldsymbol y}}

\newcommand{\deltav}{{\boldsymbol \delta}}

\newcommand{\thetav}{{\boldsymbol \theta}}

\newcommand{\E}{\mathbb{E}}

\newcommand{\Lcal}{\mathcal{L}}

\newcommand{\Dcal}{\mathcal{D}}

\newcommand{\approach}{\textsc{AT-BMC}}

\newif\ifsubscripterror\subscripterrortrue
\ifsubscripterror

\else

\fi

\usepackage{microtype}

\title{Unifying Model Explainability and Robustness for Joint Text Classification and Rationale Extraction}

\author{
    Dongfang Li \textsuperscript{\rm 1},
    Baotian Hu \textsuperscript{\rm 1}\footnote{Corresponding authors},
    Qingcai Chen \textsuperscript{\rm 1,2}\footnotemark[1],
    Tujie Xu \textsuperscript{\rm 1},
    Jingcong Tao \textsuperscript{\rm 1},
    and Yunan Zhang \textsuperscript{\rm 1},
    
}
\affiliations{

    \textsuperscript{\rm 1} Harbin Institute of Technology (Shenzhen), Shenzhen, China \\
    \textsuperscript{\rm 2} Peng Cheng Laboratory, Shenzhen, China \\
    crazyofapple@gmail.com, hubaotian@hit.edu.cn, qingcai.chen@hit.edu.cn
}

\usepackage{bibentry}

\begin{document}

\maketitle

\begin{abstract}
Recent works have shown  explainability and robustness are two crucial ingredients of trustworthy and reliable text classification.  However, previous works usually address one of two aspects: i) how to extract accurate rationales for explainability while being beneficial to prediction; ii) how to make the predictive model robust to different types of adversarial attacks.  Intuitively, a model that produces helpful explanations should be more robust against adversarial attacks, because we cannot trust the model that outputs explanations but changes its prediction under small perturbations.
To this end, we propose a joint classification and rationale extraction model named \approach{}. It includes two key mechanisms: mixed Adversarial Training (AT) is designed to use various perturbations in discrete and embedding space to improve the model's robustness, and Boundary Match Constraint (BMC) helps to locate rationales more precisely with the guidance of boundary information. 
Performances on benchmark datasets demonstrate that the proposed \approach{} outperforms baselines on both classification and rationale extraction by a large margin. Robustness analysis shows that the proposed \approach{} decreases the attack success rate effectively by up to 69\%. The empirical results indicate that there are connections between robust models and better explanations. 
\end{abstract}

\section{Introduction}

\begin{table}[ht]
\includegraphics[width=\columnwidth]{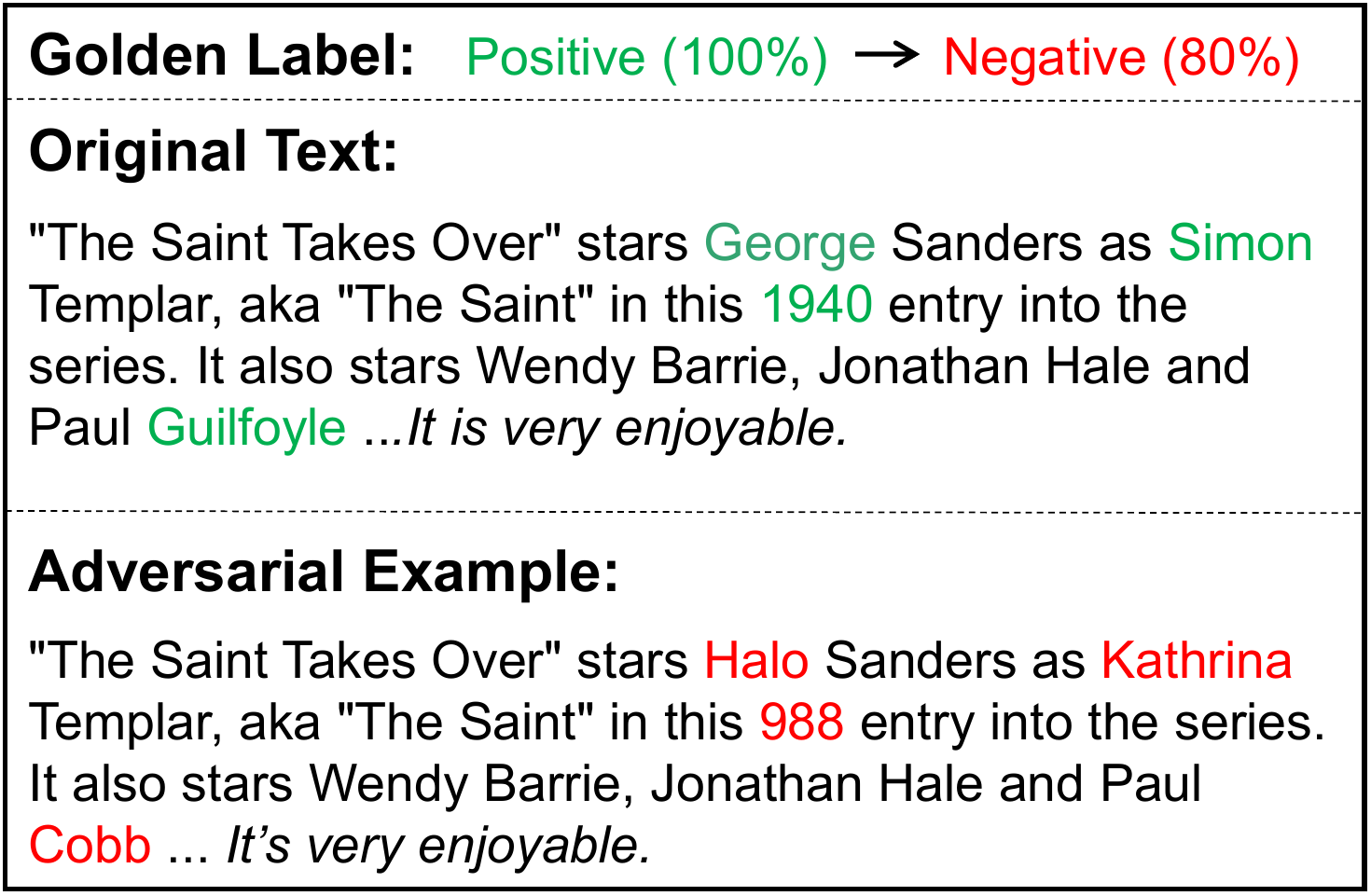}
\vspace{-2mm}
\caption{We test the pre-trained BERT-base movie review classifier using one adversarial example generated by \textsc{CheckList}~\cite{ribeiro-etal-2020-beyond}. The general meaning of the text remains unchanged. However, the predictions of the model change from \emph{positive} to \emph{negative}, while the associated rationales which enable users to verify the predictions quickly do not change. Human-labeled rationales are shown in the \textit{italic} font.}
\label{fig::example}
\vspace{-4mm}
\end{table}
Neural models have demonstrated their superior ability on text classification task, especially when based on pre-trained language models (PLMs)~\cite{bert,roberta}. However, they are more like black boxes compared to traditional machine learning methods such as logistic regression and decision tree. It is notoriously difficult to understand why neural models produced particular predictions~\cite{samek2017explainable,rudin2019stop}. 
One practical approach is to extract prediction's rationales from input~\cite{mythos-lipton-2016,camburu_e_snli_2019,thorne2019generating}. The rationales can be defined as text snippets or subsets of the input text. The assumption is that a correct prediction can be made from the rationale alone~\cite{lei2016rationalizing,bastings2019interpretable,deyoung2019eraser}. 
In other words, rationales should be sufficient to support the model's prediction. 
Our work also falls into this scope, which aims to achieve better prediction performance and model explainability by extracting prediction closely-related rationales.

Previous works proposed to use the pipeline approach where task prediction is performed in two steps: the explanation phase and the subsequent prediction phase~\cite{lei2016rationalizing}. The challenge is to attain superior task performance conditioned on the extractive rationale. 
Since most works that employ this framework tend to rely solely on task labels, they sample rationales from input in the explanation phase. For example, these models simulate the intractable sampling step by proposing optimization procedures based on reinforcement learning approaches~\cite{lei2016rationalizing,yoon2018:invase} and reparameterization techniques~\cite{bastings2019interpretable,Latcinnik2020ExplainingQA}, which may be sensitive to hyperparameters and requires complicated training process~\cite{jain2020learning}. 
Instead, we optimize the joint likelihood of class labels and extractive rationales for the input examples.
Although it is a relatively straightforward way to optimize the explanation phase models, there are at least two challenges for this task. 
Firstly, previous works are vulnerable to different types of adversarial attack. For example, a classifier suffers from defending against the labeling-preserving adversaries as shown in Table~\ref{fig::example}. If adding small perturbations to the input modifies the model's prediction, we cannot trust the explanations output from the model. We further analyze existing methods suffer from text attacks by using robustness test, which performs model-agnostic attacks on the trained classification model (\S \ref{sec:rob}). 
Secondly, the explicit boundary information is ignored, leading to inaccurate extraction. For example, ``interesting'' and ``inspiring'' are boundaries of the rationale for the text ``this film is interesting and inspiring.'', while ``is'' and ``.'' are general tokens whose representation is different from emotion words.
Besides, models that use rationales to train explanation phase models do not consider the supervision signal from task~\cite{deyoung2019eraser}. 

To address these challenges, we propose a joint classification and rationale extraction framework \approach{} where \emph{task prediction} and \emph{rationale extraction} are learned jointly with mixed Adversarial Training (AT) and Boundary Match Constraint (BMC).
Firstly, we apply perturbation in both the discrete text space and the embedding space to improve both the generalization and robustness of the model. On the one hand, we generate adversarial examples at word-level while preserving the rationale unchanged. The perturbations also maintain prediction invariance. On the other hand, our adversarial training in the embedding space refines the standard adversarial training~\cite{madry2017towards} in computation efficiency and training smoothness. Secondly, we consider matching constraints by modeling both the boundary positions, which allows the model to further focus on the boundary-relevant regions. The main idea of boundary constraint is to make the sequence labeling model to consider the boundary information when locating entities. By matching a predicted start index of a rational span with its corresponding end index, the global sequence labeling information is fused with local region-aware information.
In addition, we condition the extraction models on the classification label through label embedding. 
We conduct extensive experiments on two benchmark datasets (i.e., \emph{Movie Reviews} and \emph{MultiRC}).
The experimental results demonstrate that \approach{} outperforms the competitive baselines on classification and rationale extraction by a large margin. Robustness analysis further shows that \approach{} can effectively improve the robustness of the models where the attack success rate decreases from 96\% to 27\% under strong adversarial attacks. The code is available at \url{https://github.com/crazyofapple/AT-BMC}.

The contributions of this paper are summarized as follows: 
\begin{itemize}
	\item Different from previous methods such as pipelines, we propose \approach{} for joint classification and rationale extraction, which allows two parts to improve each other. We also show that our approach can be applied in the context of only limited annotated examples.  
    \item To the best of our knowledge, this is the first work that considers explainability and robustness both in one text classification model. As a step towards understanding the connection between explainability and robustness, we provide evidence that robust models lead to better rationales.
\end{itemize}

\section{Related Work}
\paragraph{Rationale Extraction}
The task aims to extract snippets that can support prediction in input sequences. These text snippets allows humans to verify the correctness of predictions quickly~\cite{zaidan2008modeling,zhang2016rationale,ross:17:right,deyoung2019eraser}. For example, \citet{paranjape-etal-2020-information} leverage the information bottleneck principle to extract rationales of desired conciseness. \citet{sha2021aaai} propose a selector-predictor method to squeeze information from the predictor to guide the selector in extracting the rationales. Unlike post-hoc methods~\cite{LIME,shap} where explanation does not come directly with the prediction, we focus on joint extraction of rationales and task predictions in this paper.

\paragraph{Adversarial Training} 
Adversarial training is used to improve the generalization ability and robustness of the model and has proven effective in various tasks~\cite{goodfellow2014explaining,madry2017towards,ribeiro-etal-2020-beyond}.
Some previous works use text perturbation by changing discrete inputs to generate adversarial training samples, and then the model is retrained again to improve the ability to defend some attacks~\cite{jialiang2017,wang2018naacl,michel2019naacl}.
For example, CLARE~\cite{li2020contextualized} modifies the inputs by using pre-trained masked language models in a context-aware manner.
These data augmentation methods assume that generating textual adversarial examples by sophisticated word or phrase perturbations would not change the labels. However, one limitation is that it is hard to enumerate all text manipulations~\cite{DBLP:conf/acl/ZangQYLZLS20}. The other type of adversarial training is to add gradient-based perturbations in the embedding space. Recent works have shown that this method achieves performance improvement with pre-trained language models~\cite{jiang2019smart,liu2020alum,cheng2020posterior}. For example, virtual adversarial training~\cite{zhu2019freelb} does not generate explicit adversarial examples. Instead, it samples random perturbations from the $\epsilon$-sphere surrounding the input and uses continuous optimization methods (e.g., smoothness regularization) to train the model. In this paper, we use mixed adversarial training for models to have the best of both worlds. 

\paragraph{Connecting Explainability with Adversarial Robustness} Previous work has shown that neural networks are easy to be attacked~\cite{textbugger,pwws,textfooler}, which naturally brings about the question of whether the application of interpretable technology to explain the predictive behavior of the model will be affected by the attack~\cite{DBLP:journals/corr/abs-1806-08049}. Some previous works~\cite{DBLP:conf/eccv/AugustinM020,DBLP:conf/kdd/DattaFLLSW21} empirically observed that robust models can be more explainable in computer vision. And though some recent studies~\cite{DBLP:conf/icml/EtmannLMS19,DBLP:conf/icml/MoshkovitzYC21} have focused on linking explainability and adversarial robustness, there is no explicit statement about existing models have both two properties. On the other hand, our goal is to focus on understanding the connection between the two in text classification tasks, and we hope it sheds light on the future development of such methods in natural language processing tasks.
\section{Method}

\subsection{Text Classification with Rationale Extraction}
The aim of this paper is to design a model that can yield accurate predictions and provide closely-related extractive rationales (i.e., supporting evidence) as potential reasons for predictions. 
Taking the sentiment classification as an example, for the text ``\emph{titanic is so close to being the perfect movie...}", the predictive label of it is \emph{positive}, and one of rationales for this prediction is ``\emph{so close to being the perfect movie}".
Therefore, text classification with rationale extraction can be formalized as follows. Given a sequence of words as input, namely $\mathbf{x}=\left[x_{1}, \cdots, x_{L}\right]$, where $L$ is the length of the sequence and $x_{i}$ denotes the $i$-th word. The goal is to infer the task label $\hat{y}$ and to assign each word $x_i$ with a boolean label denoted as $\hat{e}_i \in\{0,1\}$, where $\hat{e}_i = 1$ indicates word $i$ is a part of the rationale. We denote the rationale of the sequence as $\mathbf{\hat{e}}\in\{0,1\}^{L}$. The corresponding golden label is denoted as $y$ and human-labeled rationale is denoted as $\mathbf{e}$, both of them is used for training. 
Here, rationales are sequences of words and hence a potential rationale is a sub-sequence of the input sequence. Note that one text sample may contain multiple non-overlapping sub-sequences as rationale spans. 
\begin{figure}[hpt]
\centering
\resizebox{\linewidth}{!}{
\includegraphics[scale=1.0]{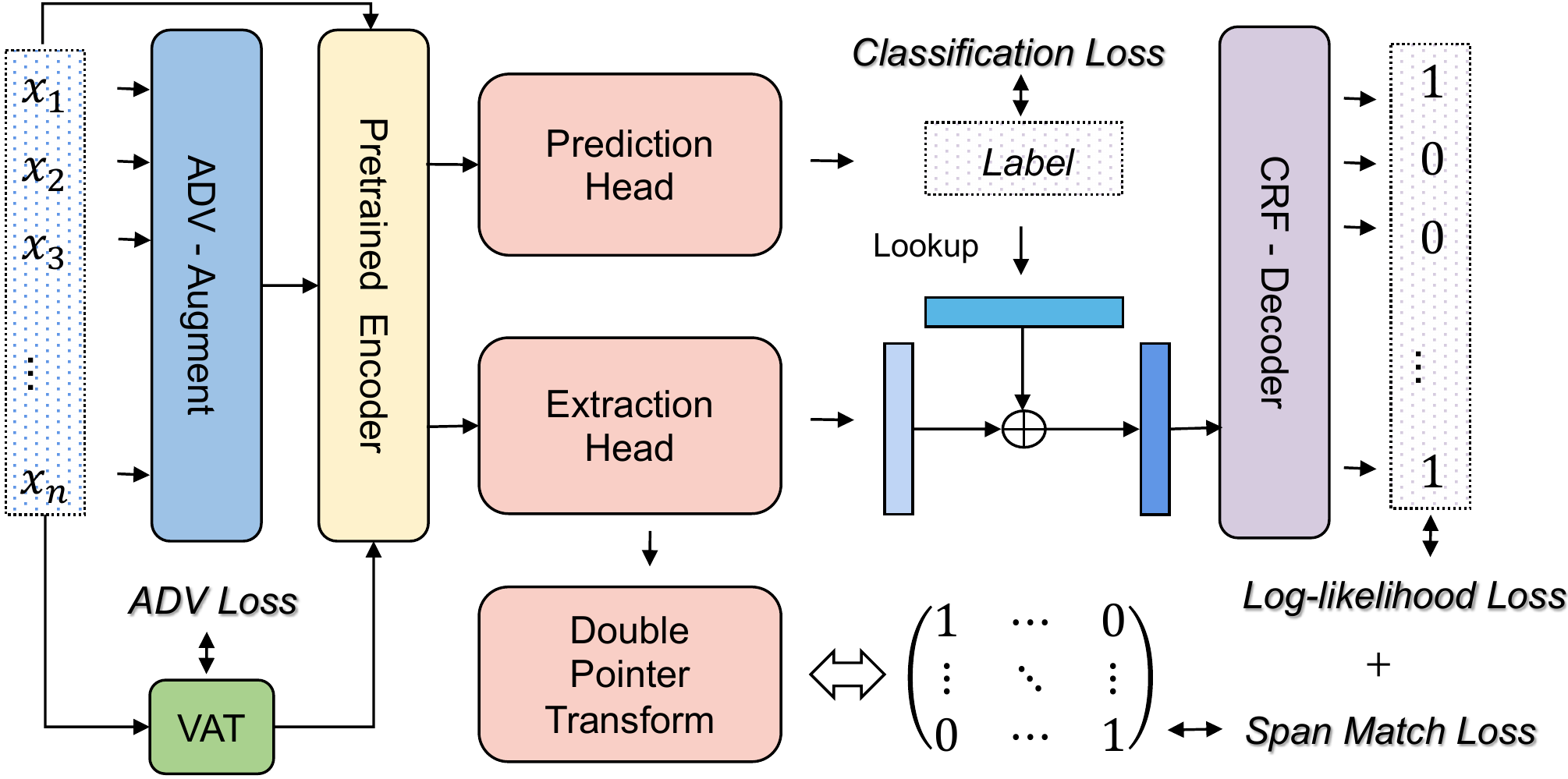}
}
\caption{Overall architecture of the proposed \approach{} method for joint classification and rationale extraction with mixed adversarial training and boundary match constraint.}
\label{fig:model}
\end{figure}
\subsection{Overall Framework}

We propose to construct a model that consists of an extraction network $g$ and a prediction network $f$, given the training data consisting of $n$ points $\mathcal{D} = \{(\mathbf{x}_1, y_1)... (\mathbf{x}_n, y_n)\}$. The prediction network $f$ first maps the text sequence  $\mathbf{x}_i$ to the task output $\hat{y}_i$. At the same time, the input $\mathbf{x}_i$ 
is also fed into rationale extraction network $g$ to output supporting evidence $\mathbf{e}_i$.

The basic scheme follows an multi-task learning (MTL) framework~\cite{caruana1997multitask} with two tasks --  (1) rationale extraction and (2) the actual prediction task.
We adopt the shared encoder architecture of MTL where both the tasks share the same encoder $\mathbf{enc}(\cdot)$ but different decoders. Formally, the conditional likelihood of the output labels and evidence, given the input, can be written as:   
\begin{equation}
L = \sum_{i=1}^n \log p(y_i, \mathbf{e}_i | \mathbf{x}_i).
\end{equation}
Note that given rationale data, our objective is to learn $\mathbf{enc}(\cdot)$, $f(\cdot)$, and $g(\cdot)$ that predict both the task labels $y_i$ and rationale labels $\mathbf{e}_i$ given $\mathbf{x}_i$. We assume that only $m$ points have evidence annotations $\mathbf{e}$.
Therefore, we can factorize the likelihood as follows:
\begin{align}
L &= \sum_{i=1}^n \log\ p(y_i | \mathbf{x}_i) \ + \sum_{j=1}^m \log \ p(\mathbf{e}_j | \hat{y}_j, \mathbf{x}_j).
\end{align}
Consequently, the predicted label and the text sequence $\mathbf{x}_i$ are fed into the extraction network $g$ to generate the evidence labels $\mathbf{\hat{e}}_i$. The overall architecture of our approach is presented in Figure~\ref{fig:model}.
Since we can optimize the classification objective for all $n$ instances and the extraction objective for $m$ instances, it allows us to train the model when only part of samples have human-labeled rationales.

Concretely, we first encode the input into hidden representation by using pre-trained language models~\cite{bert,roberta} as the shared encoder. Then we use a linear classifier to model $p_\theta(y|\mathbf{x})$ with the cross entropy loss $\mathcal{L}_{classify}$ and a linear-chain CRF~\cite{lafferty2001conditional} to model $p_\phi(\mathbf{e}|\mathbf{x}, y; \theta)$ with negative loglikehood $\mathcal{L}_{extract}$. The outputs are predicted label by the linear classifier and rationale spans generated by the CRF decoder.
Inspired by~\citet{wang-etal-2018-joint-embedding}, we also condition the extraction model on the predicted label output from the classification model. We implement this by using an embedding lookup layer, and add the label embedding to each token representation of encoder. Moreover, applying label-specific embedding can help to validate  different behaviour of the rationales via changing the $\hat{y}_i$.

\subsection{Mixed Adversarial Training}
Since the searching space of adversarial attacks is large and marked rationales is limited, we perform discrete adversarial attack based data augmentation on the samples with rationale. 
By introducing the word-level perturbed versions of existing samples, we can recursively reuse  augmentation to significantly expand the training data set. For simplicity, the validation here only considers adding one new textual edit for each sample.
In addition, considering the label-preserving of rationales, perturbations only include those parts of the sentence that are outside of the rationales. 

Specifically, we change text by using the linguistic transformations following the \textsc{CheckList} behavioral testing~\cite{ribeiro-etal-2020-beyond}. We use 4 invariance test by using \textit{TextAttack}~\cite{morris2020textattack}, including name replacement, position replacement, number change, and contraction/expansion of named entities. The invariance test is to apply perturbations that retain the label to the input and to expect the model predictions to remain unchanged. The more detail descriptions of each transformation are: (a) name replacement refers to the conversion of the input by replacing the name of the recognized name entity; (b) location conversion refers to the conversion of the recognized location entity of a sentence to another location given the location dictionary; (c) number change refers to the recognition of the number in the sentence to return the sentence with the changed number; and (d) the last transformation refers to the expansion or contraction of the identified name entity combination. 
All replacement words (from pre-defined named entity dictionary) have the same part-of-speech as the original word tagged by \textit{flair}~\cite{akbik2019flair}. Here, the percentage of words to replace per augmented example is set to $0.2$.

Apart from applying perturbation to the input text directly, we also leverage adversarial training operated on the embedding space as an effective regularization to improve the shared encoder $\mathbf{enc}(\cdot)$ generalization and reduce robust error. It aims to minimize the following objective:
\begin{align} \label{eqn:outer_min}
    \min_{\boldsymbol \theta} \mathbb{E}_{(\boldsymbol x,\boldsymbol y)\sim \mathcal{D}} \{ \mathcal{L}_{adv} + \mathcal{L}_{classify}; \boldsymbol \theta\}\,,
\end{align}
\begin{align} \label{eqn:adv_loss}
     \mathcal{L}_{adv} =  \alpha\cdot\mathcal{L}_{pat} + \beta\cdot\mathcal{L}_{kl} \,,
\end{align}
where ${\Lcal_{classify}=L(f_{\thetav}(\xv),\yv)}$ is the cross-entropy loss on original data, $\Lcal_{pat}$ is the perturbations-based adversarial training loss (PAT), and $\Lcal_{kl}$ is a smoothing adversarial regularization term. $\alpha$, $\beta$ is a hyperparameter. We define the PAT loss as: 
\begin{align} \label{eqn:inner_max}
    \Lcal_{pat} (\thetav) = \max_{||\deltav||\leq \epsilon} L(f_{\thetav}(\xv+\deltav),\yv)\,,
\end{align}
where $L$ is the cross-entropy loss on adversarial embeddings and $\deltav$ is the perturbation. We constrain $\deltav$ by using Frobenius norm bounded by $\epsilon$. The outer minimization in Eqn. (\ref{eqn:outer_min}) can be dealt with SGD for optimization. And the inner maximization in Eqn. (\ref{eqn:inner_max}) can be solved reliably by projected gradient descent (PGD)~\cite{madry2017towards}. It is a standard method for large-scale constrained optimization, which takes the following step with step-size $\eta$ at each iteration:
\begin{equation} 
    \deltav_{t+1} = \Pi_{||\deltav||\leq \epsilon} (\deltav_{t}+\eta g(\deltav_{t})/ ||g(\deltav_{t}) ||_F)\,,
\end{equation}
where $g(\deltav_{t}) = \nabla_{\deltav}L(f_{\thetav}(\xv+\deltav),\yv)$ is the gradient of the loss w.r.t. $\deltav$, and $\Pi_{||\deltav||\leq \epsilon}$ performs a projection onto the $\epsilon$-ball.
To further regularize the trade-off between standard objective and adversarial objective, we consider label smoothness in the embedding neighborhood by using term $\Lcal_{kl} (\thetav)$, which is defined as:
\begin{equation} \label{eqn:adv_regularization}
    \begin{split}
    \Lcal_{kl} (\thetav) &= \max_{||\deltav||\leq \epsilon} L_{kl}(f_{\thetav}(\xv+\deltav),f_{\thetav}(\xv))\,,
    \end{split}
\end{equation}
where $\small{L_{kl}(p,q) = [\mbox{KL}(p||q)+ \mbox{KL}(q||p)}]/2$, $p$, $q$ denote the two probability distributions, and $\mbox{KL}(\cdot)$ denotes the Kullback-Leibler divergence with temperature equals $1.0$. 
In contrast to Eqn. (\ref{eqn:inner_max})  which promotes adversarial attacks that retain labels, Eqn. (\ref{eqn:adv_regularization}) further asserts that the confidence level of the prediction should also be similar on the probability vector, which is characterized by the simplex form $\Delta$ where its dimension equals the number of classes.

Compared with standard training, $K$-step PGD requires $K$ forward-backward passes through the network, which is computationally expensive. 
Besides, only the last step of perturbation is used for model parameter update after $K$-step.
We follow free adversarial training framework in FreeLB~\cite{zhu2019freelb} to perform multiple PGD iterations to construct the adversarial embedding and iterate the cumulative parameter gradient $\nabla_{\thetav}\mathcal{L}$ in each iteration. Afterward, the model parameters $\thetav$ are updated one at a time with the accumulated gradients effectively, by virtually creating one batch that is $K$ times larger than sampled mini-batch. 
For convenience, we provide the details of adversarial training on embedding space in Algorithm~\ref{alg:freemat}.

\setlength{\textfloatsep}{16pt}
\begin{algorithm}[t!]
    \small
	\caption{
		Embedding-level Adversarial Training Algorithm
	}
	\label{alg:freemat}
	\begin{algorithmic}[1]
		\Require Training samples $\Dcal$, perturbation bound $\epsilon$, learning rate $\tau$, ascent steps $K$, ascent step size $\eta$, the total number of epochs $T$, the variance of the random initialization $\sigma^{2}$.
		\State Initialize $\thetav$
		\For{epoch $= 1 \ldots T$}
		\For{each batch $B$ sampled from $D$}
        \State $\delta \sim \mathcal{N}\left(0, \sigma^{2} I\right)$
        \State $\gv_0^{\theta} \gets 0$
		\For{$t =1 \ldots K$}
		\State Accumulate gradient of $\thetav$ given $\deltav_{t-1}$:
	    \State $\small{\gv_t^{\theta} \gets \gv_{t-1}^{\theta} + \frac{1}{K}\E_{B}[\nabla_\thetav (\Lcal_{adv} + \Lcal_{classify})]}$
		\State Update the perturbation $\deltav$ via gradient ascend:
        \State $\gv_{adv}^{\deltav} \gets  \nabla_{\deltav} \, \mathcal{L}_{adv}$
		\State $\deltav_{t} \gets \Pi_{\lVert \deltav \rVert_F\le \epsilon}(\deltav_{t-1} + \eta \cdot \gv_{adv}^{\deltav} / \lVert \gv_{adv}^{\deltav} \rVert_F$)
		\EndFor
        \State  $\thetav \gets \thetav - \tau \gv_K^{\theta}$
		\EndFor
		\EndFor
	\end{algorithmic}
\end{algorithm}

\subsection{Boundary Match Constraint}
For rationale extraction, the start/end boundaries can be captured by CRF decoder. As the CRF learns the conditional probability of the label sequence given the observation sequence features, it can be seen as maximum log-likelihood objective function conditioned on the observation $X$. However, CRF has the limitation of occasionally generating illegal sequences of tags, as it only encourages legal transitions and penalizes illegal ones softly~\cite{crf_mask}. Hence, we propose to use a boundary constraint to encourage it to be more accurate when positioning the boundary.

The basic idea of boundary constraint is to match a predicted start index of a rational span with its corresponding end index.
Given the sequence hidden representations $\mathbf{H}$ output from $\mathbf{enc}(\cdot)$, we first predict the probability of each token as the starting indices, as follows:
\begin{equation}
    \mathbf{P}_\text{s}=\text{Softmax}(\mathbf{H} \mathbf{W}_\text{s})\in\mathbb{R}^{L\times 2}\,,
\end{equation}
where $\mathbf{W}_\text{s}\in\mathbb{R}^{d\times 2}$ is the weights to learn and $d$ is the hidden size. Each row of $\mathbf{P}_\text{s}$ presents the probability distribution of each index being the start position of an word. 
Similarly, we can calculate the end index prediction logits by using another matrix $\mathbf{W}_\text{e}$ to obtain probability matrix $\mathbf{P}_\text{e}$:
\begin{equation}
    \mathbf{P}_\text{e}=\text{Softmax}(\mathbf{H} \mathbf{W}_\text{e})\in\mathbb{R}^{L\times 2}.
\end{equation}
After that, by applying \emph{argmax} to each row of $\mathbf{P}_\text{s}$ and $\mathbf{P}_\text{e}$, we obtain the predicted indexes that might be the starting or ending positions, i.e., 
 $\hat{E}_\text{s}$  and $\hat{E}_\text{e}$:
\begin{equation}
\begin{aligned}
\mathbf{\hat{E}}_\text{s}, \mathbf{\hat{E}}_\text{e} =\{i,j~|~\text{argmax}(\mathbf{P}_\text{s}^{(i)})=1, \\ 
\text{argmax}(\mathbf{P}_\text{e}^{(j)})=1\}\,,    
\end{aligned}
\end{equation}
where $i=1,\cdots,L$, $j=1,\cdots,L$. Given any start index $i\in \mathbf{\hat{E}}_\text{s}$ and end index  $j\in \mathbf{\hat{E}}_\text{e}$, 
a binary classification model is trained to predict 
the probability that they should be matched: 
\begin{equation}
\label{equ4}
    o_{i,j}=\text{sigmoid}(\mathbf{W}\cdot[\mathbf{H}_{i}, \mathbf{H}_{j}])\,,
\end{equation}
where $\mathbf{W}\in\mathbb{R}^{1\times 2d}$ is the weights to learn. Hence, the span match loss is 
\begin{equation}
\label{eqn:match_loss}
    \mathcal{L}_{match} = - \sum_{i,j}c_{i,j}\log{o_{i,j}}\,,
\end{equation}
where $c_{i,j}$ denotes the golden labels for whether this row-column position should be matched with the start index and end index of rationale spans.
\subsection{Training}

We define the final weighted loss as follow,
\begin{equation}
\small
\mathcal{L} = \mathcal{L}_{classify} + \mathcal{L}_{extract}  + \lambda_1\mathcal{L}_{adv}  + \lambda_2\mathcal{L}_{match},
\end{equation}
where $\lambda_1, \lambda_2$ are hyper-parameters of $\mathcal{L}_{adv}$ in Eqn. (\ref{eqn:adv_loss}), $\mathcal{L}_{match}$ in Eqn. (\ref{eqn:match_loss}), respectively. 
Note that at the training time, all ground-truth labels are fed into the models to learn all components of rationale extraction network and prediction network simultaneously. The marked rationales for the label in training are the collection of input text sequences annotated by humans. The same pre-trained encoder is used by shared parameters. During inference, the model recognizes rationale spans of samples with golden human-labeled rationale for extraction evaluations.

\begin{table*}[ht]
    \centering
    \small
    \begin{tabular}{@{}lccccccccc@{}}
    \toprule
    \multirow{2}{*}{Methods} & \multicolumn{3}{c}{Movie Reviews}   & & & \multicolumn{3}{c}{MultiRC}  & \\  \cline{2-4} \cline{7-9}
    &  \multicolumn{1}{c}{Accuracy} &  & \multicolumn{1}{c}{Token F1} & & & \multicolumn{1}{c}{Accuracy} &  & \multicolumn{1}{c}{Token F1}  \\ 
    \toprule
    \begin{tabular}[c]{@{}c@{}}Pipeline approach~\cite{lehman2019inferring} \end{tabular}                          & $76.9$                                &  & $14.0$ & & & $65.5$ &  & $45.6$                             \\ 
    \begin{tabular}[c]{@{}c@{}}Information Bottleneck (IB)~\cite{paranjape2020information} \end{tabular}                          & $82.4$                                &  & $12.3$ & & & $62.1$ &  & $24.9$                             \\ 
    \begin{tabular}[c]{@{}c@{}}IB (semi-supervised, 25\%)~\cite{paranjape2020information} \end{tabular}                          & $85.4$                                 & & $18.1$ & & & $66.4$ &  & $54.0$                             \\ 
    \begin{tabular}[c]{@{}c@{}}FRESH~\cite{jain2020learning}\end{tabular}                            & $93.1$                           &  & $27.7$ & & & $66.1$ &  & $53.2$ \\ 
    \begin{tabular}[c]{@{}c@{}}Weakly- \& Semi-supervised~\cite{pruthi-etal-2020-weakly} \end{tabular}                                                                                & $93.2$                           &  & $46.3$ & & & $65.4$ &  & $47.8$ \\ \midrule
    \begin{tabular}[c]{@{}c@{}} AT-BMC (BERT-base)\end{tabular}        & $\bf94.0\pm0.31$                          &  & $\bf50.6\pm0.54$          & & & $\bf67.7\pm0.49$ &  & $\bf57.3\pm0.34$  \\
    \begin{tabular}[c]{@{}c@{}} \quad w/o label embedding\end{tabular}        & $93.7\pm0.19$                            &  & $49.8\pm0.51$           & & & $65.3\pm0.35$  &  & $56.2\pm0.47$  \\
    \begin{tabular}[c]{@{}c@{}} \quad w/o adversarial examples\end{tabular}        & $93.5\pm0.47$                           &  & $48.0\pm0.34$         & & & $66.6\pm0.53$  &  & $55.5\pm0.48$   \\
    \begin{tabular}[c]{@{}c@{}} \quad w/o span match loss\end{tabular}        & $93.8\pm0.18$                           &  & $47.4\pm0.45$          & & & $67.2\pm0.32$  &  & $55.7\pm0.43$   \\
    \begin{tabular}[c]{@{}c@{}} \quad w/o virtual adversarial training\end{tabular}        & $93.6\pm0.37$                          &  & $48.1\pm0.43$         & & & $63.6\pm0.47$ &  & $55.5\pm0.51$ \\ 
    \begin{tabular}[c]{@{}c@{}} \quad w/o mixed adversarial training\end{tabular}        & $93.1\pm0.45$                          &  & $47.7\pm0.59$         & & & $63.9\pm0.41$ &  & $54.7\pm0.53$ \\ \midrule
    \begin{tabular}[c]{@{}c@{}} AT-BMC (RoBERTa-large)\end{tabular}        & $\bf95.8\pm0.43$                           &  & $\bf59.6\pm0.61$          & & & $\bf76.4\pm0.31$ &  & $\bf64.8\pm0.52$
                    \\ \bottomrule
    \end{tabular}
    \vspace{-2mm}
    \caption{Performance comparison on two text classification tasks with rationale extraction. We report test set results of AT-BMC that using different encoders (i.e., BERT-base and RoBERTa-large). Results of our AT-BMC method are averaged across $3$ different seeds. }
    \vspace{-4mm}
    \label{tab:results}
\end{table*}
\section{Experiments}
\label{sec:results}
\subsection{Dataset}
\noindent \textbf{Movie Reviews}~\cite{pruthi-etal-2020-weakly}.
This dataset includes $50$k movie reviews from IMDB dataset~\cite{maas-etal-2011-learning} and $1.8$k movie reviews with human-labeled rationales collected by~\citet{zaidan2007using}.
The labels marked by the annotator are binary sentiment indicators (i.e., \emph{Positive} and \emph{Negative}). The training set, development set, and test set consist of $26198$, $12800$, and $12800$ available examples, respectively, while the types of movies in each subset are different. Note that the samples with rationale are divided into $1200$, $300$,  and $300$ respectively. Due to the long text of the review comment, it is necessary to verify the correctness of the prediction by extracting evidence. Moreover, it is also applicable to adversarial attack scenarios.

\noindent \textbf{MultiRC}~\cite{khashabi2018looking}. The corpus comprises of multiple-choice questions and answers from various sources along with supporting evidence. This dataset concatenates each answer candidate to one question and assigns a binary label to it (i.e., whether this answer can answer the question or not). Each QA pair is associated with a related passage that is annotated with sentence-level rationales. The training set, development set, and test set consist of $24029$, $3214$, and $4848$ available examples. The rationale in this dataset contains sufficient context to allow the human to discern whether the given answer to the question is \emph{True} or \emph{False}.

\subsection{Experimental Results}
\paragraph{Evaluation Metrics}
For classification, we use classification accuracy between the predicted class label and the actual label. For rationale extraction, we report the token F1 to evaluate the quality of extraction. The micro-averaged F1 score computes at the token level between the predicted evidence spans tokens and the gold rationale tokens in terms of sets of predicted positions. 
\paragraph{Experimental Setup}
We use the BERT-base model released by Google to encode text~\cite{bert}. We also compare the performance of \approach{} which uses the pre-trained RoBERTa large model~\cite{roberta}. Our model is orthogonal to the choice of the pre-trained language model.
We use AdamW optimizer~\cite{adamw} with a batch size of $4$ for model training. The initial learning rate, the maximum sequence length, the dropout rate, the gradient accumulation steps, the training epoch and the hidden size $d$ are set to $2 \times 10^{-5}$, $512$, $0.1$, $8$, $30$, $768$ respectively. We clip the gradient norm within $1.0$. The learning parameters are selected based on the best performance on the development set. Early stopping is also applied based on model performance on the development set. Our models are trained with NVIDIA Tesla V100s (Ubuntu 18.04 LTS \& PyTorch).  We set the perturbation size $\epsilon=1\times 10^{-5}$, the step size $\eta=1\times10^{-3}$, ascent iteration step $K=2$ and the variance of normal distribution $\sigma=1\times10^{-5}$. The weight parameters $\lambda_1$, $\lambda_2$ are set to  $1.0$, $0.1$ respectively. 
The augmented adversarial examples of both datasets are $1198$ and $24007$.
\paragraph{Comparison of Baselines}
We compare our \approach{} method with following competitive methods for classification and rationale extraction in both datasets: 1) The pipeline approaches~\cite{lehman2019inferring,lei2016rationalizing} use independent parts for both extraction and prediction. These two pipeline modules are trained with classification labels and rationales labels, respectively. 2) The information bottleneck method~\cite{paranjape2020information} extracts sentence-level rationales by measurement of maximal (and minimal) mutual information with the label (and input).
3) The FRESH approach~\cite{jain2020learning} extract $k$ tokens by using attention scores for downstream classification. 4) The weakly- and semi-supervised methods~\cite{pruthi-etal-2020-weakly} present a classify-then-extract framework condition the rationale extraction on the classification. The pipeline approach uses RNNs, whereas the base model is BERT-base for IB, FRESH, and~\cite{pruthi-etal-2020-weakly}, as same as ours for a fair comparison. The performances of baselines are from reference papers. As shown in Table~\ref{tab:results}, our model improves over the previous models on both datasets. In the task of rationale extraction, \approach{} (BERT-base) and \approach{} (RoBERTa-large) improves $4.3$ and $13.3$ F1 points over the previous models on the Movie Reviews dataset. Moreover, on the MultiRC dataset, our method also improves the F1 up to $3.3$ and $10.8$ points. On the other hand, \approach{} (BERT-base) improves $0.8$ and $1.3$ in terms of accuracy respectively, which might come mainly from two main aspects: one is multi-task learning and the other is adversarial training.

\begin{table} [ht!]
  \resizebox{\linewidth}{!}{
  \begin{tabular}{lccc}
    \toprule
    \toprule
     Metrics & Baseline & \citet{pruthi-etal-2020-weakly} & AT-BMC \\
    \midrule
    Acc. (Test) & $90.00$ & $93.20$ & $93.97$ ($0.77$ $\uparrow$)\\
    Avg. words & $216.96$	& $216.96$ & $216.96$\\
    \midrule
    \midrule
    \multicolumn{3}{l}{\textbf{TextFooler}~\cite{textfooler}} & $ $ \\
    \midrule
    Acc. (Attack) & $1.00$ & $9.00$ & $22.00$ ($21.00$ $\uparrow$) \\
    SR  &	$98.89$	& $90.22$	& \textcolor{red}{$57.69$ ({$42.2$} $\downarrow$)} \\
    Avg. PW 	& $10.02$	& $10.88$	& $23.65$  \\ 
    Avg. AQ	& $741.44$	& $980.23$ & $2300.40$  \\ 
    Total time	& $1078.26$	& $1552.26$	& $3118.59$ \\ 
    \midrule
    \multicolumn{3}{l}{\textbf{TextBugger}~\cite{textbugger}} & $ $ \\
    \midrule
    Acc. (Attack) & $9.00$ & $31.00$ & $38.00$ ($29.00$ $\uparrow$) \\
    SR  &	$90.00$	& $67.02$	& \textcolor{red}{$26.92$ ($63.08$ $\downarrow$)} \\
    Avg. PW  	& $31.26$	& $33.97$	& $41.61$  \\ 
    Avg. AQ	& $583.01$	& $834.19$ & $1497.04$ \\ 
    Total time	& $1480.80$	& $2098.11$	& $2337.80$ \\ 
    \midrule
    \multicolumn{3}{l}{\textbf{PWWS}~\cite{pwws}} & $ $ \\
    \midrule
    Acc. (Attack) & $3.00$ & $4.00$ & $38.00$ ($35.00$ $\uparrow$) \\
    SR  &	$96.67$	& $95.74$	& \textcolor{red}{$26.92$ ($69.75$ $\downarrow$)} \\
    Avg. PW  	& $5.87$	& $4.99$	& $17.44$ \\ 
    Avg. AQ	& $1505.1$	& $1459.83$ & $2299.5$  \\ 
    Total time	& $2005.26$	& $2125.88$	& $2356.88$ \\ 
    \bottomrule
    \bottomrule
  \end{tabular}
  }
  \caption{Comparison of baselines and AT-BMC using three adversarial attack methods. The baseline is the vanilla BERT fine-tuned on the IMDB dataset~\cite{maas-etal-2011-learning}. We compare performance in accuracy under attack, attack success rate (SR), average percentage of perturbed words in the sentence (PW), average attack queries of all examples (AQ) and total attack time. All numbers are reported on 100 test instances. $\uparrow$ ($\downarrow$) represents that the higher (lower) the better.}
  \label{tab:robust}
\end{table}
\subsection{Robustness Evaluation}
\label{sec:rob}

The robustness test is to perform model-agnostic attacks on the trained classification model. It verifies the robustness of our method to attack samples; hence we can better classify and extract evidence by which humans verify predictions. It assumes that the model input is text sequences and returns outputs that the objective function can process. In this process, the attack algorithm will find the disturbance of the input sequence that satisfies the attack target and follows a certain language restriction. In this way, the attack to models is framed as a combinatorial search problem. In order to find a series of perturbations that produce a successful adversarial example, the attacker must search through all possible conversions. We refer the reader to~\cite{morris2020textattack} for
more details.

\begin{figure}[ht!]
\centering
\includegraphics[width=0.8\columnwidth]{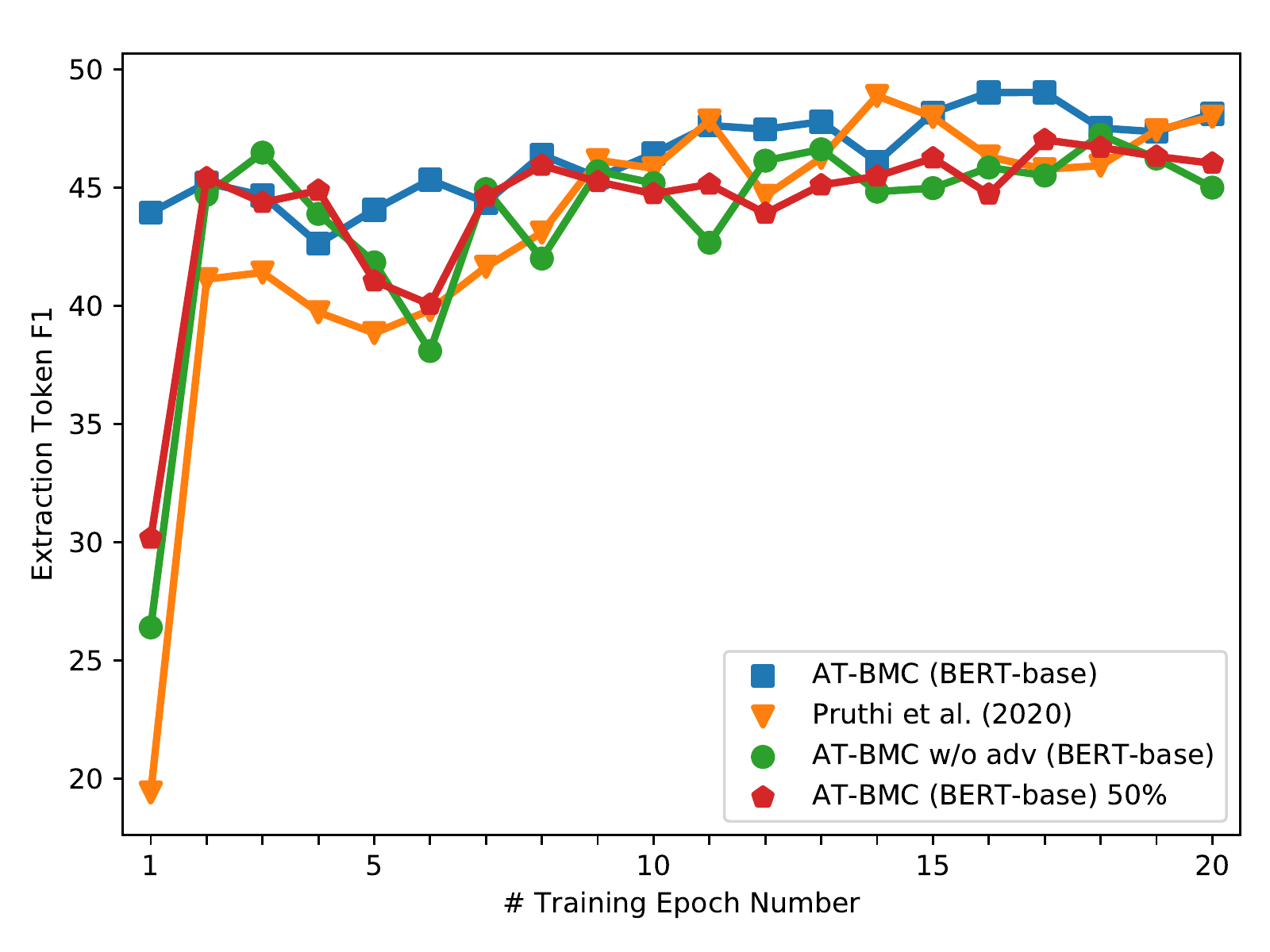}
\caption{Extraction F1 curves on the development set of Movie Reviews. Our model with BERT-base is trained on 100\%, 50\% of the fraction of rationales in the training set.}
\label{fig::we}
\end{figure}
\begin{figure}[htp]
\centering
    \subfloat[Movie Reviews]{{\includegraphics[width=4cm]{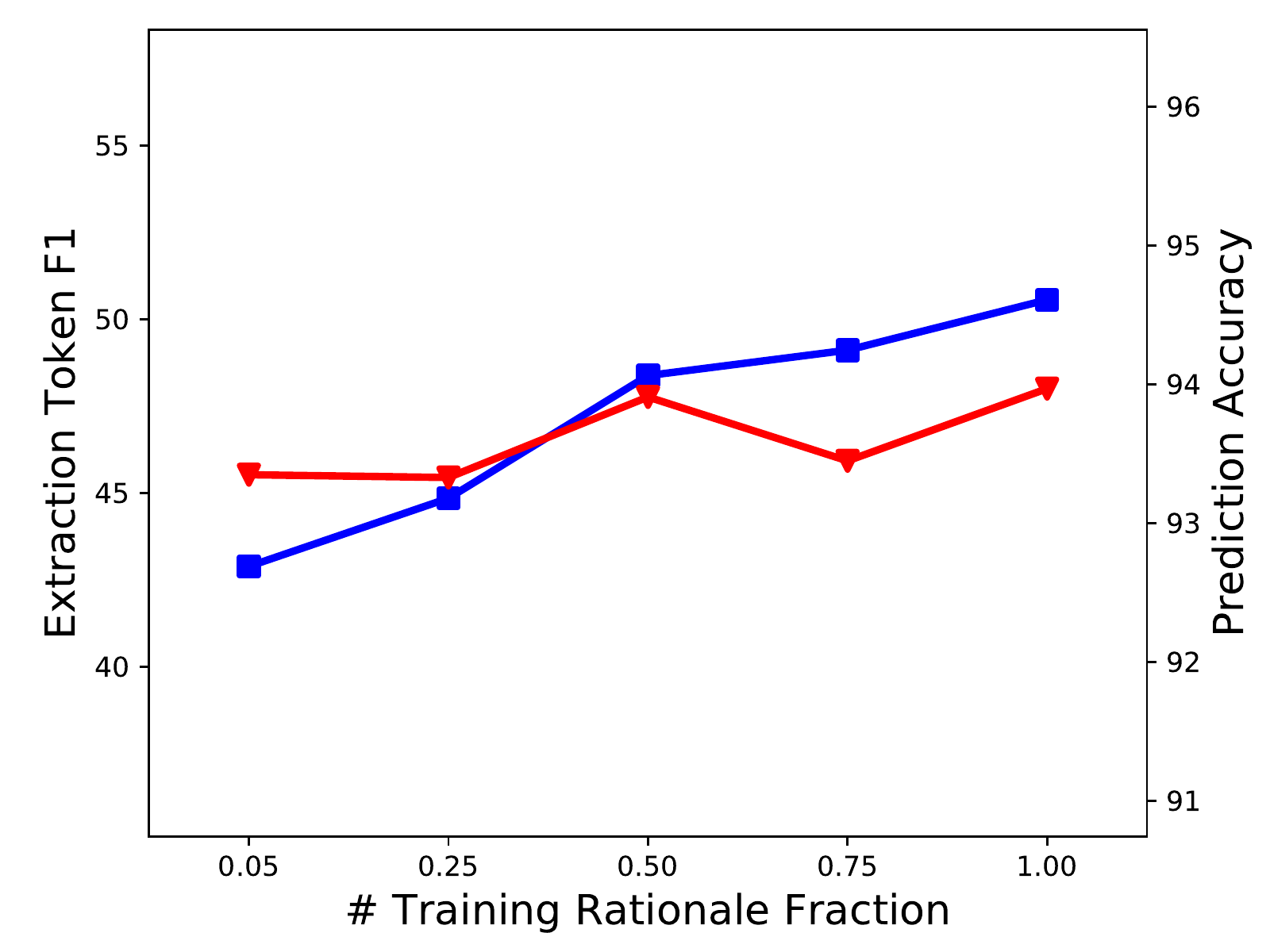}}}
    \subfloat[MultiRC]{{\includegraphics[width=4cm]{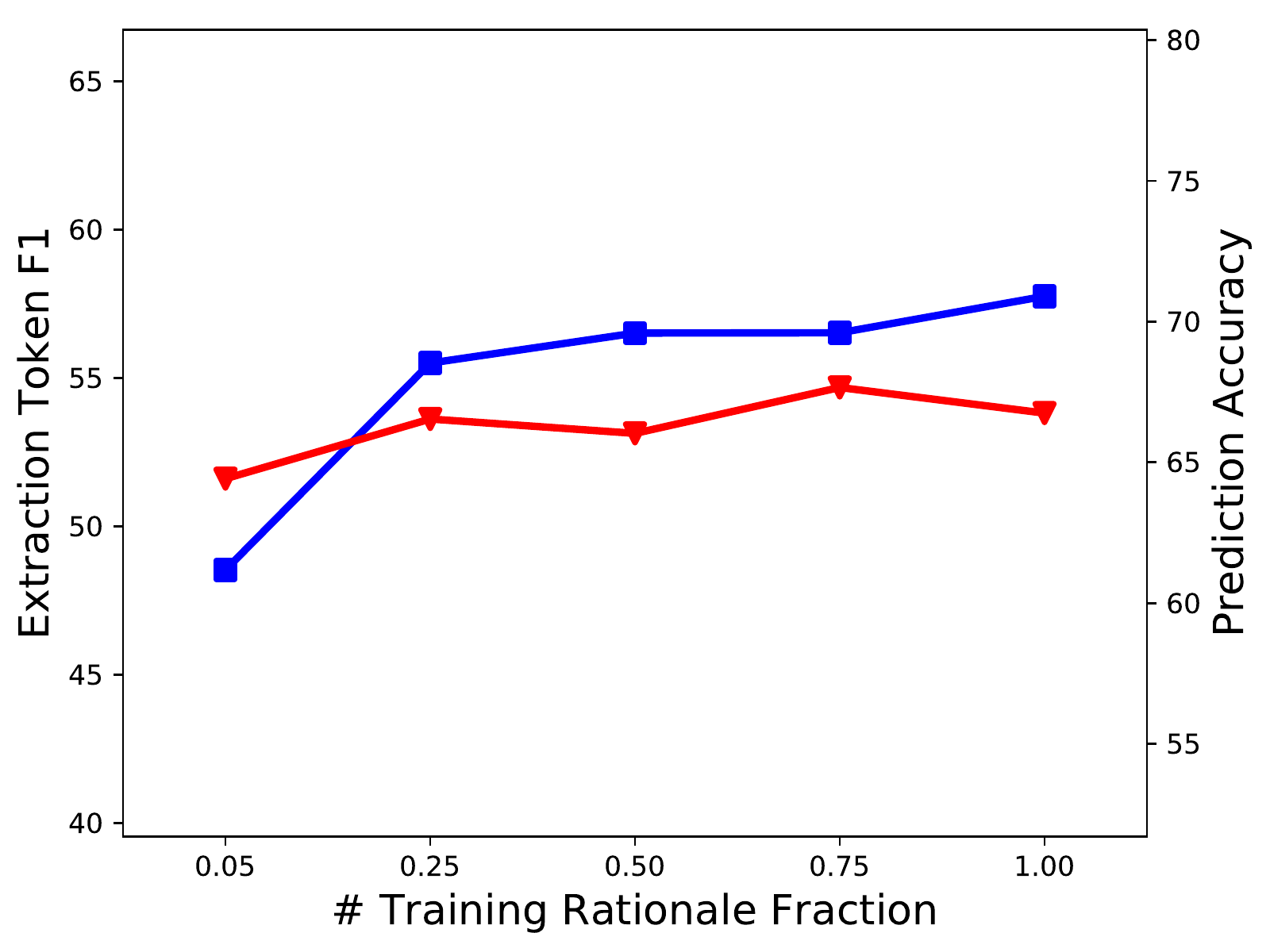}}}
\vspace{-2mm}
\caption{Comparison in task accuracy shown in red and extraction performance shown in blue as the number of rationales retained in the training set varies.}
\vspace{-4mm}
\label{fig:increasing_data}
\end{figure}
Instead of measuring robustness by \textit{interpretability robustness} where rationales should be invariant to small perturbations in the input, we consider three different attack methods (i.e., TextFooler~\cite{textfooler}, TextBugger~\cite{textbugger}, and PWWS~\cite{pwws}) to \textit{classification robustness}. In the test, we focus on the success rate of the attack. TextFooler and TextBugger use a mixture of measures (e.g., word embedding distance, part-of-speech tags match), and a word replacement mechanism is designed to attack the existing model; PWWS greedily uses word importance ranking to replace parts of sentences, where word saliency and synonym swap scores are used to calculate word importance. The results of classification models on the Movie Reviews test set are presented in Table~\ref{tab:robust}. Overall, \approach{} achieves the best performance on all metrics consistently across different attack methods. Notably, \approach{} substantially outperforms the baseline by 69.75\% success rate under PWWS attack.  We attribute this to \approach{}’s generalizability obtained by adversarial training. Interestingly, from the results in the second column, using a joint framework also seems to enhance generalizability and robustness in this domain. We also report the percentage of words being replaced for attacking as average word modification rates. Our method needs more modified attack queries with higher word modification rates under all attacks. It indicates that the model is harder to attack and hence requires more words to be replaced.
In Figure~\ref{fig::we} we compare the evaluation curves of the different methods on the development set over training time. We take the performance on the Movie Reviews development set at each evaluation checkpoint. The magnitude of change in our method is much smaller relative to other methods that did not perform adversarial training and boundary match constraint, and it converges gradually as training time changes. It illustrates that our method reinforces the robustness of the model on the development set during training, thereby making it more stable for training and less variance, which alleviates hyperparameter sensitivity and high variance in existing methods.

Our approach can also be applied in the context of which there are only limited annotated examples. As shown in Figure~\ref{fig:increasing_data}, we compare the performance of models with varying proportions of human-labeled rationales in the training set. We find that the model achieves extraction accuracy above 40 on the test set when only 5\% of examples with labeling signals. As the ratio of these marker examples increases, the performance of the model improves. Since the manual labeling of these annotations is time-consuming and labor-intensive, this might imply that our approach can stably generate reasonable interpretations without many manual annotations.

\subsection{Analysis and Discussion}
\paragraph{Ablation Study}
To study the effect of each part, we conduct ablation experiments. The results are also shown in Table~\ref{tab:results}. From the experimental results, adversarial examples, boundary match loss, and virtual adversarial training all contribute to the performance improvement of the model, each helping to improve the model performance by about two percentage points in rationale extraction. We can see that the match loss improves token F1 relatively the most on the Movie Reviews dataset, probably because after adding this term, the extraction model may focus more on local boundary information. For the original task accuracy, adversarial training can boost performance, and both adversarial data augmentation and virtual adversarial training in the embedding space can bring improvements. It illustrates that these two methods are complementary. The results also show that considering on the predicted label improves the extraction F1 by $0.8$ and $1.1$ points on both datasets.
\paragraph{Effectiveness Evaluation} Human-labeled rationales contain some implicit information that represents partial inductive bias in predictions. The interpretable AI community would like to use them as a guide for evaluating model interpretations and possibly for teaching models to make robust and reasonable decisions. Compared with the answer prediction task, existing models have relatively lower prediction results on the Token F1 metric in the rationale extraction task. Hence, we randomly sampled 50 correct examples and ask two annotators to judge the relevance between extracted rationales and labels. The inter-rater agreement coefficient between two annotators is 0.85. The relevance results of two datasets are 45 and 47, which means high scores between predicted labels and extracted rationales. We also compare the model without the boundary match constraint. The relevance results are 39 and 42. The comparison shows that by matching starting and ending tokens, the model can be more relevant to labels and aligned with human interpretation.
\paragraph{Error Analysis}
We conduct an error analysis on extraction rationales generated by our model on 50 randomly chosen examples from the development set of Movie Reviews. The major error types are  summarized as follows: 
1) the dominant type (48\%) is that the model outputs contain bags of small fragments, which overlap with human evidence. For example, some fragments are emotional expressions, such as ``\textit{perfect}'', ``\textit{some of the best}''. 2) the second type (24\%) is caused by incomplete or inadequate annotations. For example, the model outputs ``\textit{it's a very impressive film}'' and ``\textit{wonderfully presented story}'', while humans only annotate the former. It demonstrated that marked rationales do not necessarily have high comprehensiveness by including all relevant information, which aligns with the findings of~\citet{DBLP:conf/emnlp/CartonRT20}. 3) the last type (28\%) is caused by several factors, such as the text is too long and the evidence is outside the truncated text; or the prediction of the model itself is wrong. It shows that only learning from the supervision signal may be affected by annotation artifacts and variance between human-annotated rationales. And how the machine can help us correct superficial clues instead of learning could be another interesting topic.

\section{Conclusion}

In this work, we focus on how to jointly classify and provide extracted rationales, so that humans can use it to verify the correctness of the prediction. We propose a method \approach{} for jointly modeling text classification and rationale extraction using mixed adversarial training and boundary match constraint. The results on two public data sets show that our method improves the performance of the model on the task, especially with increasing extraction token F1 for rationales. 
Besides, \approach{} can remarkably decrease the attack success rate compared to the baseline under different attack methods. The results indicate that robust models lead to better extracted rationale in text classification tasks.
In the future, we will explore how to apply our model to more domains (e.g., medical and legal domains). 

\section*{Acknowledgements}
We thank the valuable feedback of Wenpeng Yin, Yuxiang Wu, Shuoran Jiang and the insightful comments and suggestions of the anonymous reviewers. This work is jointly supported by grants: Natural Science Foundation of China (Grant No. 61872113 and Grant No. 62006061), Stable Support Program for Higher Education Institutions of Shenzhen (No. GXWD20201230155427003-20200824155011001), Strategic Emerging Industry Development Special Funds of Shenzhen (No. XMHT20190108009 and No. JCYJ20200109113403826), Fundamental Research Fund of Shenzhen (No. JCYJ20190806112210067), and Tencent Group.
\bibliography{aaai22}

\end{document}